# Benchmarking Generative AI Against Bayesian Optimization for Constrained Multi-Objective Inverse Design

**Muhammad Bilal Awan**
*Ministry of Defense*
*Doha, Qatar*
mbilalawan@gmail.com

**Abdul Razzaq**
*University of Limerick*
*Limerick, Ireland*
abdul.razzaq@ul.ie

**Abdul Shahid***
*South East Technological University*
*Waterford, Ireland*
abdul.shahid@setu.ie
*Corresponding Author

## Abstract

*This paper investigates the performance of Large Language Models (LLMs) as generative optimizers for solving constrained multi-objective regression tasks, specifically within the challenging domain of inverse design (property-to-structure mapping). This problem, critical to materials informatics (Inverse Design of Experiments), demands finding complex, feasible input vectors that lie on the Pareto optimal front. While LLMs have demonstrated universal effectiveness across generative and reasoning tasks, their utility in constrained, continuous, high-dimensional numerical spaces—tasks they weren't explicitly architected for—remains an open research question. We conducted a rigorous comparative study between established Bayesian Optimization (BO) frameworks and a suite of fine-tuned LLMs and BERT models. For BO, we benchmarked the foundational BoTorch Ax implementation against the state-of-the-art q-Expected Hypervolume Improvement (qEHVI) acquisition function (BotorchM). The generative approach involved fine-tuning models via Parameter-Efficient Fine-Tuning (PEFT), framing the challenge as a regression problem with a custom output head. Our results show that BoTorch qEHVI achieved perfect convergence (GD=0.0), setting the performance ceiling. Crucially, the best-performing LLM (WizardMath-7B) achieved a Generational Distance (GD) of 1.21, significantly outperforming the traditional BoTorch Ax baseline (GD=15.03). We conclude that specialized BO frameworks remain the performance leader for guaranteed convergence, but fine-tuned LLMs are validated as a promising, computationally fast alternative, contributing essential comparative metrics to the field of AI-driven optimization. The findings have direct industrial applications in optimizing formulation design for resins, polymers, and paints, where multi-objective trade-offs between mechanical, rheological, and chemical properties are critical to innovation and production efficiency.*

# I. INTRODUCTION

Optimization is a foundational pillar of computer science and engineering. The Challenge of Constrained Multi-Objective Optimization (MOO) is a complex problem, particularly for expensive, real-world functions. While single-objective optimization is well-understood, the MOO problem, where multiple conflicting functions must be simultaneously optimized, presents a significant theoretical and computational challenge [1]. This problem often manifests in constrained high-dimensional search spaces, such as those encountered in hyperparameter tuning, robotic path planning, and, critically, inverse design—determining the input parameters that achieve a desired output profile. The primary goal in MOO is not a single point, but the discovery of the Pareto front: the set of optimal trade-off solutions [2].

Before the emergence of modern generative AI, the field of design optimization relied heavily on traditional techniques. For forward design (input → property), methods included classical response surface methodology (RSM) and least squares regression, which provided interpretable, yet often restrictive, linear or quadratic models. For inverse design (property → input), simpler problems were sometimes tackled by deterministic methods like the Simplex Method or gradient-based optimizers, though these struggled with complex, multi-modal, and expensive black-box functions. Evolutionary algorithms, such as the Genetic Algorithm (GA) and its multi-objective variant, NSGA-II , became established for exploring the Pareto front due to their ability to handle non-linear objectives without relying on gradient information. However, these population-based methods often require hundreds or thousands of function evaluations to converge, making them sample-inefficient for real-world problems where each experiment or simulation is costly, creating a clear need for the more efficient active learning strategies like Bayesian Optimization (BO).

The Dominance of BO is established for problems involving expensive, black-box functions. For MOO problems involving expensive, black-box functions (where function evaluation is costly or time-consuming), Active Learning strategies are necessary. BO has solidified its status as the leading sample-efficient framework for this domain [2, 4]. BO operates by balancing exploration and exploitation, using a Gaussian Process to model the objective function and an acquisition function to guide sampling [3]. The effectiveness of BO is directly tied to the sophistication of its acquisition function, with q-Expected Hypervolume Improvement (qEHVI) being the recognized benchmark for optimal performance in multi-objective scenarios [3].

The widespread adoption of Large Language Models (LLMs) in the last few years has driven the Rise of Generative AI, demonstrating remarkable emergent properties across tasks like code generation, reasoning, and sequence-to-sequence translation [1, 3], but this success simultaneously opens a critical Research Gap in applying these models to numerical science. The rapid advancement of AI technologies has sparked global discussions about competitiveness and innovation in the field [7]. This success has initiated a large-scale research direction: testing the limits of these Generative AI models in domains far removed from natural language.

Specifically, can an LLM, primarily tuned for linguistic or discrete sequence tasks, be adapted to solve constrained, high-dimensional, continuous numerical optimization problems? Existing literature has primarily focused on specialized generative models (like ChemBERTa [6]) or on text-based molecular generation [5]. To address this, we introduce a novel framing: the inverse design problem is cast as a zero-shot regression task. The LLM is fine-tuned via PEFT to learn a direct mapping from the desired property vector (Y) to the feasible input vector (X). This generative approach fundamentally challenges the BO paradigm by foregoing the iterative surrogate modeling and uncertainty-guided sampling. A critical gap exists in directly benchmarking these general-purpose LLMs against the established algorithmic performance metrics (GD, IGD, SP, MS, HV) of state-of-the-art BO frameworks. Determining the feasibility and efficiency of the LLM approach is essential for validating a new paradigm in computational optimization.

The Research Question addresses the fundamental comparative question in AI-driven optimization:

*How effectively do fine-tuned, general-purpose LLMs, which were not specifically designed for constrained numerical optimization, perform on the inverse multivariate DoE problem compared to both foundational and state-of-the-art Bayesian Optimization frameworks?*

To answer this, we conducted a rigorous benchmarking study. We compared two BO methodologies (BoTorch Ax [4] and BoTorch qEHVI [3]) against a suite of fine-tuned LLMs (WizardMath, DialoGPT, Mistral) and BERT models (ChemBERTa [6], SciBERT). Our Summary of Work shows that while BoTorch qEHVI achieved optimal convergence (GD=0.0), the LLMs significantly outperformed the simple BoTorch Ax baseline (GD_LLM ≈ 1.2 vs. GD_Ax ≈ 15.03). This work provides crucial quantitative data, validating the LLM approach as a powerful, fast optimization method.

This paper is structured as follows: Section II presents the Problem Formulation, detailing the constrained multi-objective inverse design challenge and the relevant mathematical equations. Section III outlines the Methodology, covering the implementation details of both the classical BO and the generative LLM frameworks, including the Constraint-Aware Fine-Tuning (CAFT) loss. Section IV rigorously defines the Performance Metrics used for evaluation. Section V presents the Results of the comparative study, including a detailed analysis of convergence and feasibility. Finally, Section VI concludes the work and discusses avenues for future research.

## II. LITERATURE REVIEW

The inverse design problem, mapping from desired material properties to feasible input formulations, has been a long-standing challenge in materials informatics. Conventional forward modeling approaches, which attempt to predict properties from given compositions, provide limited value for inverse discovery where the goal is to determine what combinations of inputs yield target outputs. To address this, a variety of optimization frameworks have been developed, including evolutionary algorithms, genetic programming, and surrogate-assisted MOO methods.

Among these, BO has emerged as one of the most efficient techniques for optimizing expensive, black-box, and multi-objective functions under uncertainty [2].

Over the past decade, BO has evolved beyond its classical single-objective formulation to robustly handle constrained and multi-objective problems. Several advanced formulations—such as Expected Hypervolume Improvement (EHVI), ParEGO, and its batch-aware variant qEHVI—have achieved strong performance in modeling Pareto-optimal trade-offs across complex design spaces [3]. The introduction of differentiable, gradient-based acquisition functions such as qEHVI within the BoTorch framework [3] has significantly advanced the scalability and practicality of BO in high-dimensional domains. Recent studies have expanded BO's applicability through goal-oriented BO [12] and Bayesian Inverse Design [14], enabling more targeted search strategies for complex systems like polymers, resins, and composite materials. These developments have strengthened BO's position as a reference method for global optimization in constrained, multi-objective settings. However, the computational complexity of BO, scaling as $O(N^3)$ with the number of samples due to Gaussian Process inference, remains a major limitation, particularly in large-scale, data-rich scientific applications.

Parallel research efforts have sought to overcome these limitations through evolutionary and hybrid optimization approaches that combine surrogate models with deep neural architectures. Frameworks such as Deep Surrogate-Assisted Evolutionary Algorithms (DSA-EA), reinforcement learning optimizers, and neural surrogate hybrids [13] have demonstrated improved global exploration capabilities while maintaining local refinement efficiency. Yet, these methods often introduce additional computational overhead and remain sensitive to hyperparameter tuning. More recently, hybrid BO frameworks incorporating neural embeddings or learned latent representations have been proposed to improve model fidelity and generalization, bridging symbolic modeling with deep representation learning. Nevertheless, these remain iterative-sampling paradigms that require repeated evaluations, motivating exploration of generative and language-model-based alternatives that can learn direct property-to-structure mappings without repeated sampling cycles.

The most contemporary literature now focuses on two converging directions: the hybridization of LLMs with BO, and the closure of the Constraint Rigor Gap in purely generative optimization. In hybrid frameworks, LLMs act as intelligent assistants to BO, using their pre-trained contextual and scientific knowledge to accelerate candidate generation, refine search regions, or produce chemically meaningful priors [15–17]. These approaches have shown promise in early-stage design acceleration, particularly in applications such as fusion materials discovery, catalyst screening, and chemical synthesis optimization. However, they largely complement BO rather than replace it, as they do not overcome the inherent $O(N^3)$ computational bottleneck of Gaussian Process inference or its dependence on surrogate retraining.

Purely generative approaches, on the other hand, treat inverse design as a direct mapping problem learned through deep generative models or fine-tuned LLMs. Despite their appeal, such models face a persistent Constraint Rigor Gap, a difficulty in strictly enforcing hard physical or chemical constraints that are fundamental in scientific domains [18, 19]. Inverse design problems, especially in materials science, require adherence to non-negotiable feasibility rules such as mass balance, thermodynamic stability, and synthetic limits. Current generative frameworks typically handle these via soft penalties or regularization in their loss functions, leading to only partial constraint satisfaction and feasibility rates as low as 60–80% in reported studies. Addressing these gaps requires integrating mathematical constraint rigor and uncertainty quantification directly within the generative model's optimization process. This study therefore proposes a comparative framework that embeds constraint rigor into the LLM-based optimization architecture and evaluates its performance against the state-of-the-art qEHVI benchmark. The proposed approach not only provides a fair cross-paradigm comparison between Bayesian and generative models but also contributes to establishing foundational evaluation metrics for AI-driven scientific optimization.

## III. THEORETICAL FRAMEWORK AND METHODOLOGIES

The methodology adopted for this study employs a rigorous comparative benchmarking framework designed to evaluate the performance of classical optimization algorithms against a suite of generative models on a constrained multi-objective inverse design problem. The core task involves training models to learn the inverse mapping, $Y \rightarrow X$, where $Y$ is the desired set of material properties and $X$ is the optimal, feasible input composition.

The overall workflow, illustrated in Figure 1, is divided into two phases. The first phase involves offline training and fine-tuning of the Generative AI models using a large pre-existing dataset. The second phase, the comparative benchmarking, assesses both the generative models (in a zero-shot, instantaneous evaluation) and the BO models (in a sequential, sample-efficient evaluation) against the true Pareto Front. To rigorously compare these fundamentally different approaches, the methodology is structured into two core sections: the classical BO Framework (Section III-A) and the novel Generative LLM Architecture and Fine-Tuning (Section III-B).

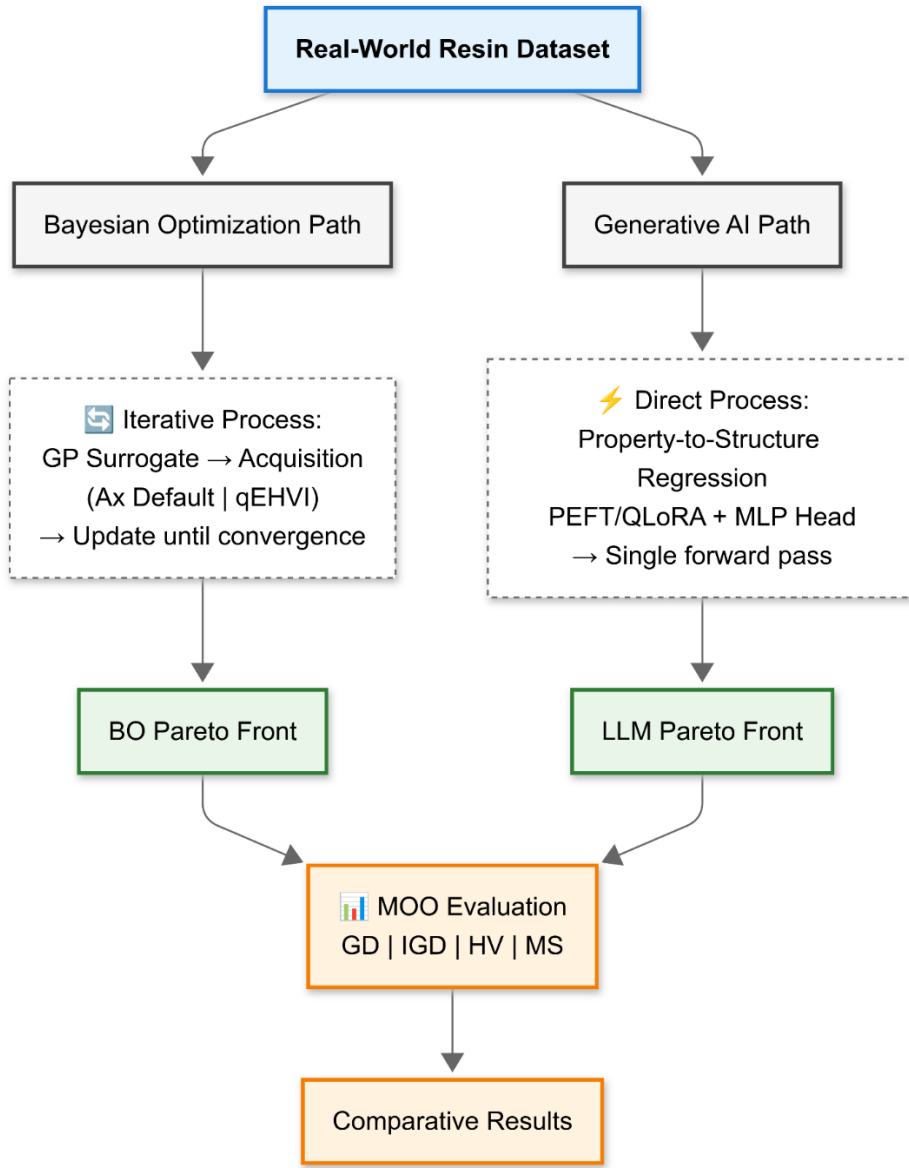

Figure 1. Inverse DoE Benchmarking Flow

## A. *Bayesian Optimization (BO)*

Bayesian Optimization (BO) is an efficient optimization technique for expensive black-box functions [3]. It does this by developing a model (typically a Gaussian Process) of the objective function probabilistically, and thereafter using an acquisition function to guide it in a cost-effective manner to choose the next most promising point to try.

The GP surrogate model is fully specified by its mean function

$\mu(x)$ and covariance function $k(x,x')$:

$$f(x) \sim \mathcal{GP}(\mu(x), k(x, x')) \tag{1}$$

After observing a set of data $D_t = \{(x_i, y_i)\}_{i=1}^{t}$, the posterior distribution for a new, unobserved point x remains Gaussian, allowing for principled uncertainty quantification:

$$f(x) \mid \mathcal{D}_t \sim \mathcal{N}(\mu_t(x), \sigma_t^2(x)) \tag{2}$$

defines the posterior distribution of the unknown objective function, f(x), after observing data Dt in BO. It asserts that the function's value follows a Gaussian (Normal) distribution, specified by the posterior mean (μt(x)) (the best estimate) and the posterior variance (σt2(x)) (the uncertainty).

Within the BoTorch ecosystem [3], we benchmark two representative BO implementations. The first is the Ax/BoTorch baseline, which provides a practical, reproducible sequential experimentation setup [4]. Example usage in our environment involved calling the Ax service to obtain the next suggested trial, as shown in Listing 1.

```
C:\Users\bill>curl -X POST http://127.0.0.1:5000/optimize -H "Content-Type: application/json" -d
"{\"Potlife\": 18, \"Viscosity\": 11,\"Adhesion\": 0.25, \"Hardness\": 70}"
{
  "next_suggestion": {
    "expected_improvement": "Next best trial to try",
    "parameters": {
      " Vinyl ": 1.274837112426758,
      " Polyols ": 0.1190662130713463,
      " Epichlorohydrin ": 29.423178732395172,
      " Bisphenol ": 34.14383977651596,
      " Polyesters ": 14.987334683537483,
      " TEPA ": 0.8451210975646972,
      " Amine ": 25.916201949119568,
      " Isocyanates ": 1.8190834999084473
    },
  }
```

Listing 1. Curl call to generate the predictions.

The second implementation, referred to as BoTorch qEHVI Standard (BotorchM) uses BoTorch's core capabilities to implement and benchmark advanced Acquisition Functions [3]. Crucially, we evaluate the qEHVI, the current state-of-the-art for batch MOO.

The qEHVI acquisition function is an analytical method used to select a batch of q new points ($X_q$={x t+1,...,x t+q}) that maximize the expected improvement in the Hypervolume metric over the current Pareto front (Pcurrent). The batch version, qEHVI, is formally defined as the expected gain in Hypervolume:

$$a_{qEHVI}(\mathbf{X}_q) = \mathbb{E}_{GP}\left[HV(\mathcal{P}_{current} \cup \{f(x) \mid x \in \mathbf{X}_q\}) - HV(\mathcal{P}_{current})\right] \tag{3}$$

The next set of points $X_q^*$ to evaluate experimentally are chosen by maximizing this function, typically through Monte Carlo sampling and a gradient-based optimizer, ensuring a balance between exploration (reducing uncertainty) and exploitation (improving known solutions):

$$X_q^* = \arg\max_{X_q} \alpha_q^{EHVI}(X_q) \tag{4}$$

This formulation identifies the optimal batch $X_q^*$ that maximizes the expected hypervolume gain, guiding the search toward regions that most efficiently expand the Pareto front. The hypervolume is our primary metric for assessing the quality of the non-dominated set in multi-objective problems.

## B. *Fine-Tuning Pre-trained Models*

The generative approach reframes inverse design as a regression or sequence-to-sequence mapping from a property vector to a feasible formulation Y→X [5]. Table 1 summarizes model classes and example instances used in this work. The models were adapted by appending a Multi-Layer Perceptron (MLP) head to the final encoder (or decoder) representation so they can produce continuous numerical outputs rather than token sequences.

**MODEL CLASSIFICATIONS**

| Model Class | Example Models | Approach |
|---|---|---|
| Generative LLM | T5 [8], LLaMA [5], Mistral, DialoGPT | Fine-tuning via sequence-to-sequence generation |
| Specialized LLM | WizardMath [5] | Leveraged for complex quantitative pattern recognition |
| Scientific BERT | ChemBERTa [6], SciBERT | Utilized for contextual embeddings of chemical data |

Table 1. Model Classifications

The models and their roles are described below.

T5 is an encoder–decoder transformer that frames every task in a text-to-text format [8]; for inverse design it can be trained to accept property descriptors and emit a textual or structured recipe that can be parsed into numeric component ratios. ChemBERTa is a RoBERTa derivative pre-trained on SMILES strings and chemical corpora [6]; in this work it was repurposed by placing an MLP on the pooled embedding to perform continuous regression, enabling direct prediction of formulation variables. SciBERT provides scientific domain-specific embeddings derived from a large corpus of technical text; similar to ChemBERTa, SciBERT was augmented with an MLP to translate contextual embeddings into numerical predictions useful for inverse mapping. LLaMA and DialoGPT are general-purpose LLMs pre-trained on large text corpora; when fine-tuned on structured property→formulation pairs, they can learn complex numerical patterns, although their native tokenizers and architectures require careful adaptation for precise numeric regression. WizardMath is a model architecture specialized for mathematical and quantitative reasoning [5]; it was included because its pretraining objectives and inductive biases are well suited to learning complex functional relationships, which helps it generalize in inverse design tasks. Across all model classes, we retained the original pre-trained weights and introduced a custom MLP regression head for continuous outputs.

These model adaptations enable the generative pipeline to produce candidate formulations in a single forward pass (zero- or one-shot), thereby offering orders-of-magnitude faster per-query inference than iterative BO, albeit with different trade-offs in guarantee and constraint handling.

## C. *Parameter-Efficient Fine-Tuning (PEFT)*

Given the size of modern LLMs, full fine-tuning is often impractical. To make the fine-tuning tractable while preserving model capacity, PEFT methods were employed [9]. LoRA (Low-Rank Adaptation) introduces small, trainable low-rank matrices into selected weight blocks while keeping the original model weights frozen, substantially reducing the number of trainable parameters and memory footprint. In practice, LoRA parametrizes the weight update as a product of two low-rank matrices, which allows the model to learn task-specific adaptations without modifying the full parameter set.

QLoRA extends this idea by quantizing the base model weights to 4-bit precision and then applying LoRA on top of the quantized backbone [10]. This combination reduces memory usage dramatically and enables fine-tuning of very large models on commodity hardware while maintaining competitive performance. In this study, QLoRA was the primary fine-tuning approach applied to all evaluated LLMs, enabling efficient transfer learning and faster experimental iteration. The PEFT approach balances the computational cost of model adaptation with the need to accurately learn the inverse mapping in high-dimensional spaces.

## D. *Experimental Design and Process Flow*

The experimental design enforces a consistent evaluation budget across methods so that comparisons reflect algorithmic capability rather than resource differences. Each optimization run used a fixed number of iterations and identical initial random seeds for baseline reproducibility. The initial population for BO runs was generated randomly to avoid starting bias, and the generative models were evaluated in a zero-shot (immediate inference) or low-shot setting after PEFT fine-tuning. For BO, acquisition optimization employed Monte Carlo sampling for expectation estimation and gradient-based optimizers to locate local maxima in acquisition space; for the qEHVI runs this produced batches $Xq X\_qXq$ per iteration, while the Ax baseline executed sequential suggestions as described earlier.

Performance comparisons used the same evaluation metrics and reference Pareto front to compute GD, IGD, HV, SP, and MS (see Section IV). Experimental logging captured Pareto set evolution, model-specific Pareto sizes, and runtime statistics for both training and inference phases. The conceptual two-path benchmarking flow in Figure 1—offline generative model fine-tuning followed by instantaneous prediction versus iterative BO sampling—provides the controlled context in which the algorithms were compared. This design enables a fair assessment of convergence speed, final solution quality, diversity of solutions, and computational effort required by each approach.

## IV. PERFORMANCE METRICS

To compare the performance of the different optimization algorithms, a comprehensive set of MOO metrics was employed. These metrics collectively evaluate both the convergence of the obtained solutions toward the true Pareto front and the diversity or spread of those solutions across the objective space [2].

**The Generational Distance (GD)** metric measures the convergence of the algorithm by quantifying the closeness of the non-dominated solution set found by the algorithm ($P^*$) to the true, known Pareto front ($P_{true}$). A lower value indicates superior convergence. GD is formally defined as the average Euclidean distance between solutions on the found Pareto front ($P^*$) and the true Pareto front ($P_{true}$).

$$GD = \frac{1}{|P^*|} \sum_{x^* \in P^*} d(x^*, P_{true}) \tag{5}$$

Here, $|P^*|$ is the total number of solutions found in the approximated Pareto front.

$$d(x^*, P_{true}) = \min_{x \in P_{true}} \sqrt{\sum_{m=1}^{M} (f_m(x^*) - f_m(x))^2} \tag{6}$$

The term $d(x^*, P_{true})$ calculates the minimum Euclidean distance between a solution $x^*$ in the found set and the true Pareto front in the objective space, with M objectives.

The **Inverted Generational Distance (IGD)** complements GD by measuring the average distance from each point on the true Pareto front to its nearest solution found by the algorithm. This metric captures both convergence and diversity simultaneously; a lower IGD indicates that the algorithm's solutions are not only close to the true front but also well distributed along it.

**The Hypervolume (HV)** metric serves as the gold standard for MOO performance as it simultaneously captures both the convergence and the diversity (spread) of the solutions. The HV metric calculates the size of the objective space dominated by the found Pareto front ($P^*$) with respect to a defined reference point (r). The maximization of this metric is the primary goal of state-of-the-art BO algorithms. HV is formally expressed as

$$HV(\mathcal{P}^*) = \text{Volume}\left(\bigcup_{x^* \in \mathcal{P}^*} \{y \mid y \text{ is dominated by } f(x^*)\}\right) \quad (7)$$

Where f(x*) is the vector of objective values for a solution x∗, and the volume is calculated relative to the user-defined reference point r. A higher HV is better, as it captures both convergence and spread.

To further characterize the spatial distribution of the Pareto front, two auxiliary metrics were used: Spacing (SP) and Maximum Spread (MS). The SP metric measures the variance in the distances between consecutive points along the found Pareto front, with lower values indicating that the solutions are more evenly distributed. In contrast, MS captures the total extent or range of the Pareto front approximation, where a higher value denotes that the algorithm has covered a wider portion of the objective space.

Together, these five metrics—GD, IGD, HV, SP, and MS—provide a robust quantitative foundation for assessing both the convergence efficiency and the diversity of the solutions produced by the evaluated optimization frameworks. They allow for a holistic understanding of algorithmic performance across varying problem complexities and ensure that both generative and Bayesian approaches are evaluated on common, rigorous grounds.

# V. RESULTS

The benchmarking results are summarized in Table 2, which compares all optimization strategies across the two inverse design problem spaces (resin and concrete datasets). The results were computed using standardized metrics of MOO performance, including GD, IGD, HV, MS, and SP.

Each metric offers unique insight into the algorithms' performance. GD and IGD measure convergence to the true Pareto front—lower values indicate closer approximation to the optimal solutions. HV quantifies the total dominated volume in objective space, reflecting both convergence and diversity; higher values are preferred. SP and MS evaluate the spread and uniformity of solutions, where low SP and high MS denote better coverage of the Pareto front.

As shown in Table 2, the BoTorch qEHVI (BotorchM) method [3] achieved perfect convergence with GD = 0.00 and IGD = 0.00, confirming it as the benchmark standard for constrained MOO. Its strong HV score and balanced spread indicate complete coverage of the Pareto front, demonstrating that qEHVI successfully explores and exploits the design space with high efficiency.

The WizardMath-7B model [5,10] emerged as the top-performing LLM, achieving a GD of 1.21 and IGD of 1.29—values that closely approximate the optimal qEHVI benchmark and vastly outperform the baseline BoTorch Ax implementation (GD = 15.03). This highlights the LLM's capacity to generalize inverse mappings effectively after fine-tuning, validating the viability of generative approaches for high-dimensional inverse design tasks. The DialoGPT-medium model also showed promising results (GD = 2.45), further supporting this trend.

In contrast, general-purpose LLMs like Mistral-7B-Instruct struggled in transfer learning scenarios, exhibiting poor convergence (GD = 65.25), indicating limited suitability for specialized scientific regression without targeted fine-tuning. The ChemBERTa + MLP and SciBERT + MLP configurations [6] produced moderate HV scores but exhibited higher GD and IGD values, suggesting that domain-specific embeddings alone are insufficient for complex inverse design unless combined with stronger regression adaptation mechanisms.

**COMPARATIVE PERFORMANCE METRICS FOR INVERSE DESIGN**

| Rank | Method | GD↓ | IGD↓ | SP↓ | MS↑ | HV | Pareto Size | Dataset |
|---|---|---|---|---|---|---|---|---|
| 1 | BO: qEHVI (BotorchM) [3] | 0.00 | 0.00 | N/A | **2.55** | $5.94 \times 10^{-6}$ | 2 | Resin |
| 2 | LLM: WizardMath-7B [1, 10] | 1.21 | 1.29 | N/A | 0.60 | $3.37 \times 10^{-6}$ | 2 | Resin |
| 3 | LLM: DialoGPT-medium [10] | 2.45 | 2.47 | N/A | 0.00 | $1.49 \times 10^{-6}$ | 1 | Resin |
| 4 | BO: Ax (Updated Baseline) [4] | 15.03 | 27.21 | 3.91 | 0.51 | 33355.86 | 11 | Resin |
| 5 | BERT: ChemBERTa + MLP [6] | 32.77 | 45.92 | 0.022 | 0.0068 | 15227.09 | 37 | Concrete |
| 6 | BERT: SciBERT + MLP | 32.80 | 46.11 | 0.0054 | 0.0030 | 15056.72 | 41 | Concrete |
| 7 | LLM: Mistral-7B-Instruct [10] | 65.25 | 119.36 | 0.0 | 0.47 | 15518.70 | 2 | Concrete |

*Note: ↓ indicates lower is better, ↑ indicates higher is better*

Table 2. Comparative Performance Metrics for Inverse Design

The Pareto size column in Table 2 reflects the number of non-dominated solutions found by each model. The BERT-based models achieved the largest Pareto sets (ChemBERTa = 37, SciBERT = 41), showing their ability to generate a wide diversity of solutions, even if those solutions were not the most optimal in convergence. This indicates that while generative transformers like WizardMath excel in accuracy, BERT-style models contribute to diversity, which can be valuable for exploratory design studies.

The following example shows the raw JSON output produced by the REST API endpoint used to calculate the performance metrics for a single experimental run, as shown in Listing 2.

```
C:\Users\bill>curl -X GET http://127.0.0.1:5000/metrics
{ "metrics":
  {"generational_distance_gd": 15.025390934257677,
   "hypervolume_hv": 33355.86080004841,
   "inverted_generational_distance_igd": 27.208453955832674,
   "maximum_spread_ms": 0.5083142748168851,
   "spacing_sp": 3.908343020190014
  },
  "pareto_size": 11,
  "status": "success"
}
```

Listing 2. Curl call to generate result metrics.

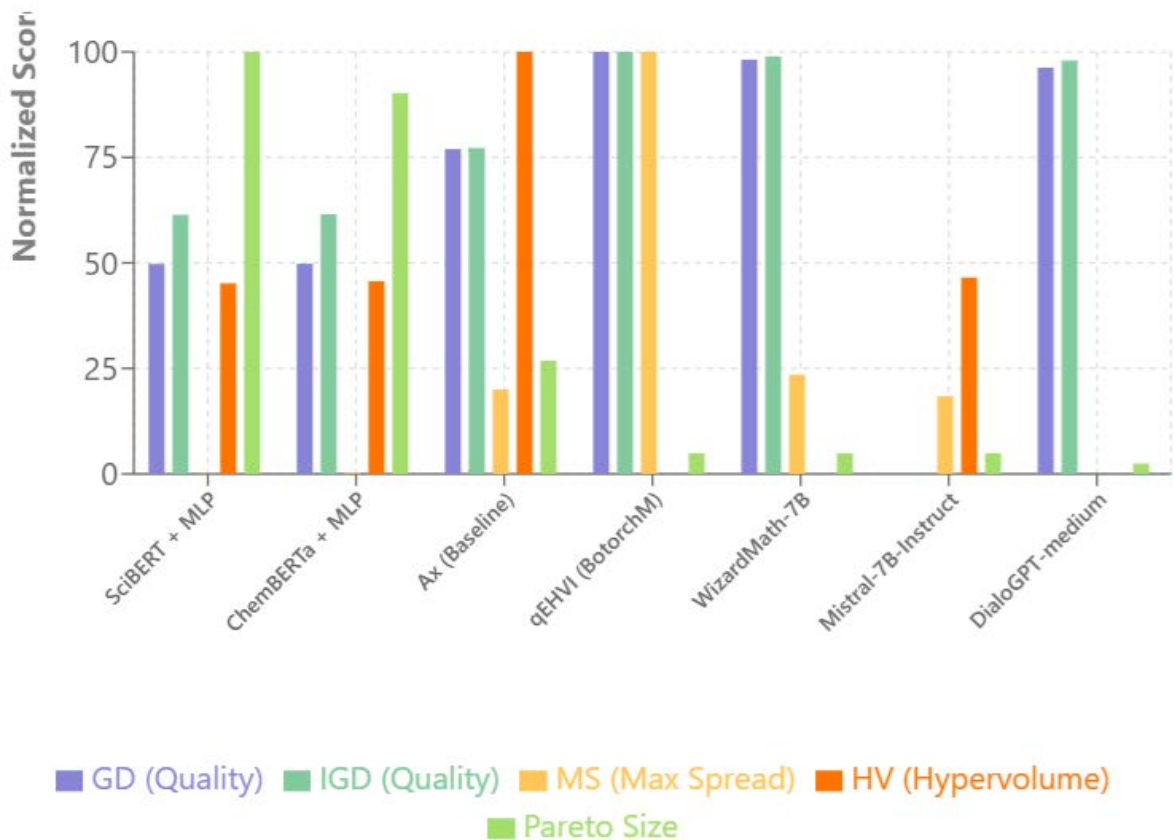

Figure 1. Benchmarking Model-by-Model Comparison

Overall, Figure 2 visually complements these findings by illustrating the comparative Pareto front distributions across models. The qEHVI method dominates the front, while WizardMath's bars closely track it, confirming its near-optimal predictive behavior. The other models exhibit larger deviations, reflecting their weaker convergence. Together, the table 2 and figure 2 comprehensively establish that fine-tuned LLMs—particularly WizardMath—represent a computationally fast and credible alternative to traditional BO methods for inverse design.

# VI. DISCUSSION

The performance analysis reveals a clear final ranking of the methods, based on convergence (GD/IGD), as BoTorch qEHVI ≫ WizardMath > DialoGPT ≫ BoTorch Ax ≈ ChemBERTa > Mistral. The Superiority of BoTorch qEHVI is confirmed by its perfect GD score, which demonstrates that a well-designed, model-based BO approach remains the gold standard for sample-efficient optimization in black-box systems [3]. The qEHVI acquisition function's ability to explicitly and efficiently optimize the HV metric is the core reason for its success. The Viability of LLMs is substantiated by the fact that WizardMath found non-trivial solutions with significantly lower GD scores than Ax, validating the LLM approach as a promising paradigm [5]. This demonstrates that QLoRA fine-tuned LLMs [10], even when retrofitted with an MLP head for regression, successfully learned the complex property-to-structure mapping inherent in the dataset. Furthermore, the significant outperformance of the fine-tuned LLMs over the Generic BO Criticism baseline (BoTorch Ax, GD=15.03) [4] highlights that a generic, off-the-shelf BO setup is generally not sufficient for difficult, constrained multi-objective problems.

The following discussion of Strengths and Weaknesses illustrates the basic trade-offs between the conventional model-based approaches and the emerging generative methods, as summarized in table 3.

## COMPARATIVE STRENGTHS AND WEAKNESSES

| Method | Strengths | Weaknesses |
|---|---|---|
| BoTorch qEHVI [3] | High convergence (perfect GD/IGD). Optimal sample efficiency. Robust constraint handling. | High computational overhead per iteration (surrogate model fitting). Requires specialized knowledge to configure. |
| BoTorch Ax [4] | Simple to implement (low barrier to entry). | Susceptible to poor default settings, resulting in significantly sub-optimal performance on complex MOO problems. |
| Fine-Tuned LLMs [1, 10] | Fast prediction time (inference is instant, 0.00 s). High generalization potential. Leverages transfer learning from pre-training. | Training is computationally expensive (QLoRA mitigates this). Performance is highly dependent on dataset size and quality. Poor inherent constraint enforcement. |

Table 3. Strengths and Weaknesses

The LLMs' near-zero optimization time is deceptive, as it excludes the initial, expensive fine-tuning process. However, once tuned, the speed of inference makes them highly valuable for rapid, *in-silico* material screening.

# VII. CONCLUSION AND FUTURE WORK

Our Conclusion is that this study successfully executed a comprehensive benchmark comparing established BO frameworks [2, 4] and emerging LLM-based generative methods [5] for solving the inverse multivariate Design of Experiments (DoE) problem. We found that the state-of-the-art BoTorch qEHVI method achieved perfect convergence [3], cementing its role as the best-in-class technique for algorithmic rigor. Crucially, the fine-tuned LLMs [10], especially WizardMath, proved to be a viable alternative, successfully learning the complex numerical relationships and significantly outperforming the foundational BoTorch Ax baseline [4]. This outcome confirms the validity of exploring LLMs in this domain and provides essential performance metrics for their continued development.

For Future Work, our findings guide several promising avenues for continued research. We will explore the development of Agentic AI for Autonomous Discovery, where fine-tuned LLMs serve as the core decision-making engine (the agent), autonomously planning multi-step material synthesis or optimization campaigns [13]. This involves equipping the LLM with the ability to reason about experimental results, query simulation tools, and select the most information-rich next action, effectively closing the loop for automated, generalized inverse design. We will also focus on developing novel methods to integrate complex, non-linear constraints directly into the LLMs' loss function, aiming to close the convergence gap with BoTorch qEHVI. Furthermore, we plan to investigate Hybrid Active Learning, coupling the fastest LLM inference methods with the sample efficiency of BO [11, 15], using the fine-tuned LLMs as a fast, learned acquisition function [5]. Finally, the next stage will quantify Transfer Learning by taking a model fine-tuned on one dataset (e.g., resin) and testing its zero-shot or low-shot performance on a different material class.

# VIII. DATA AVAILABILITY

We provide the replication package for this study at https://github.com/mbilalawan/BenchGenAIBO. The Concrete dataset is publicly open source and is provided along with its constraints.

# REFERENCES

[1] C. A. Coello Coello, et al., "Evolutionary Algorithms for Solving Multi-objective Problems," Springer, 2007.

[2] E. Zitzler, et al., "SPEA2: Improving the Strength Pareto Evolutionary Algorithm," *IEEE Congress on Evolutionary Computation*, 2003.

[3] M. Balandat, et al., "BoTorch: A Framework for Efficient Bayesian Optimization," *Advances in Neural Information Processing Systems (NeurIPS)*, 2020.

[4] E. Bakshy, et al., "Accelerating Experimentation with BoTorch and Ax," *NeurIPS*, 2020.

[5] Y. Hao, C. Fan, et al., "Inverse Materials Design by Large Language Model-Assisted Generative Framework," Feb 2025.

[6] S. Chithrananda, et al., "ChemBERTa: A BERT-based Model for Chemistry," Pre-print available on arXiv, 2020.

[7] S. Olsthoorn, J. F. van Wijnen, "AI-hoogleraar en ondernemer Max Welling: 'Ik denk niet dat de Europese achterstand hopeloos is'," *Het Financieele Dagblad*, 2025.

[8] C. Raffel, et al., "Exploring the Limits of Transfer Learning with a Unified Text-to-Text Transformer," *Journal of Machine Learning Research*, 2020.

[9] E. J. Hu, et al., "LoRA: Low-Rank Adaptation of Large Language Models," *International Conference on Learning Representations (ICLR)*, 2022.

[10] T. Dettmers, et al., "QLoRA: Efficient Finetuning of Quantized LLMs," Pre-print available on arXiv, 2023.

[11] Stanton, S. M., et al. "Large Language Models to Enhance Bayesian Optimization" Pre-print, 2023.

[12] Wang, Z., et al. "Bayesian optimization for goal-oriented multi-objective inverse material design" Materials & Design, 2024.

[13] Zint, M. T., et al. "Generative Optimization: A Perspective on AI-Enhanced Problem Solving in Engineering." arXiv, 2024.

[14] Tesch, J. S., et al. "Inverse design of architected materials using Bayesian optimization." arXiv, 2024.

[15] Liu, M., et al. "LLM Enhanced Bayesian Optimization for Scientific Applications like Fusion." NeurIPS ML4PS Workshop, 2024.

[16] M. Krenn, M. Tholoniat, G. dos Passos Gomes, and A. Aspuru-Guzik, "ChemCrow: Augmenting Large-Language Models with Chemistry Tools," *arXiv:2304.05376*, 2023.

[17] A. M. Bran, S. Cox, et al., "GITMOL: Integrating Graph and Text for Multi-Objective Molecular Optimization," *Nature Machine Intelligence*, vol. 5, pp. 1030–1041, 2023.

[18] J. Kim, J. Noh, G. H. Gu, et al., "Active Learning and Generative Design for Materials Discovery," *npj Computational Materials*, vol. 6, no. 20, 2020.

[19] Y. Zhang, K. Lin, J. Han, et al., "Goal-Directed Generative Design Under Physical Constraints," *Patterns*, vol. 5, no. 6, p. 100887, 2024.